\documentclass[10pt,twocolumn,letterpaper]{article}

\usepackage[final]{cvpr}      

\usepackage{multirow}
\usepackage{booktabs}
\usepackage[x11names]{xcolor}
\usepackage[normalem]{ulem}
\usepackage{subcaption}
\usepackage{graphicx}

\useunder{\uline}{\ul}{}

\definecolor{cvprblue}{rgb}{0.21,0.49,0.74}
\usepackage[pagebackref,breaklinks,colorlinks,allcolors=cvprblue]{hyperref}

\def\paperID{16063} 
\def\confName{CVPR}
\def\confYear{2026}

\title{No Hard Negatives Required: Concept Centric Learning Leads to Compositionality without Degrading Zero-shot Capabilities \\of Contrastive Models}

\vspace{-20mm}\author{
    Hai X. Pham\thanks{Correspondence to: $<$pham.xuan.hai@outlook.com$>$}
    \quad David T. Hoffmann
    \quad Ricardo Guerrero
    \quad Brais Martinez
    \\
    Samsung AI Center Cambridge, UK
}

\begin{document}
\maketitle

\begin{abstract}
Contrastive vision-language (V\&L) models remain a popular choice for various applications. However, several limitations have emerged, most notably the limited ability of V\&L models to learn compositional representations. Prior methods often addressed this limitation by generating custom training data to obtain hard negative samples. Hard negatives have been shown to improve performance on compositionality tasks, but are often specific to a single benchmark, do not generalize, and can cause substantial degradation of basic V\&L capabilities such as zero-shot or retrieval performance, rendering them impractical. In this work we follow a different approach. We identify two root causes that limit compositionality performance of V\&Ls: 1) Long training captions do not require a compositional representation; and 2) The final global pooling in the text and image encoders lead to a complete loss of the necessary information to learn binding in the first place. As a remedy, we propose two simple solutions: 1) We obtain short concept centric caption parts using standard NLP software and align those with the image; and 2) We introduce a parameter-free cross-modal attention-pooling to obtain concept centric visual embeddings from the image encoder. With these two changes and simple auxiliary contrastive losses, we obtain SOTA performance on standard compositionality benchmarks, while maintaining or improving strong zero-shot and retrieval capabilities. This is achieved without increasing inference cost.
We release the code for this work at \url{https://github.com/saic-fi/concept_centric_clip}.
\end{abstract}

\section{Introduction}
\label{sec:intro}

\begin{figure*}[ht]
    \centering
    \subfloat[]{
        \includegraphics[width=0.302\linewidth]{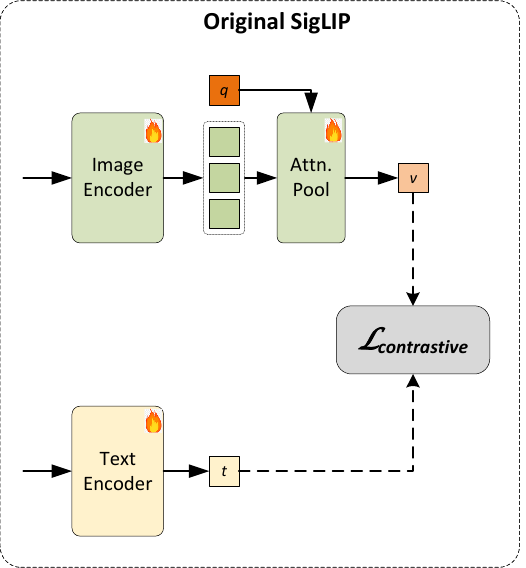}
        \label{fig:siglip}
    }
    \subfloat[]{
        \includegraphics[width=0.69\linewidth]{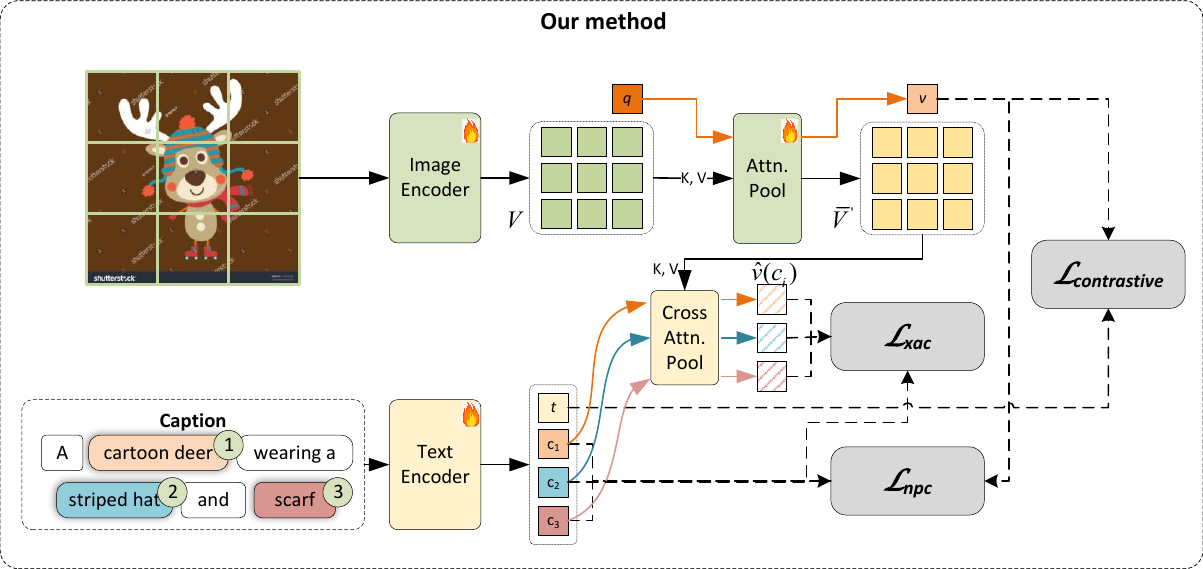}
        \label{fig:our_method}
    }
    \vspace{-0.5\baselineskip}
    \caption{\textbf{Method overview}. 
    \textbf{(a)} SigLIP uses a learnable query token in combination with an attention layer to pool the visual tokens into a single token. Aligning only global representations hampers the learning of a compositional representation.
    \textbf{(b)} Similar to SigLIP, our method aligns the global representations $v$ and $t$. To simplify learning of a compositional representation, our method extends SigLIP by first, pooling the text encoder output tokens into concept embeddings $\{c_k\}$, which are used to attention-pool concept-specific information from the visual tokens $\bar{V}$, resulting in $\hat{v}(c)$. $\hat{v}(c_k)$ and the corresponding $c_k$ are aligned using $L_{xac}$. Furthermore, the global visual representation $v$ is aligned with all $c_k$ via ${\mathcal{L}}_{\textbf{npc}}$, a multi-positive variant of SigLIP loss. Similarly, the global image and text representations $v$ and $t$ are aligned via $\mathcal{L}_{contrastive}$.
    }
    \label{fig:method_overview}
\end{figure*}

Contrastive vision-language (V\&L) models have been a cornerstone of computer vision and machine learning since the publication of CLIP~\citep{clip_openai}.
While contrastive V\&L models have achieved remarkable performance on a large variety of tasks and enabled deployment of vision models to open world settings, various limitations have been found. Most notably, they struggle when learning compositional representations, such as binding a noun to its attributes, understanding the relations between objects, or recognizing when one of multiple objects has been replaced by another object \cite{negclip_iclr23,sugarcrepe,sugarcrepe++}.
Compositionality is a core capability of vision systems and results in various failure cases if not learned. Therefore, many works have focused on ways to improve compositionality post-hoc \citep{negclip_iclr23,assouel2025object, gurung2025common, tang_emnlp23, Singh2024-cn,ce-clip-Zhang2023-vj,Doveh2023-qp,Doveh2022-yg,Patel2024-qj,sugarcrepe,lewis2022does,sugarcrepe++}, most of which \citep{Singh2024-cn,ce-clip-Zhang2023-vj,Doveh2023-qp,Doveh2022-yg,Patel2024-qj} rely on clever ways of constructing hard negatives for fine-tuning purposes. 

Previous works correctly identified one of the reasons why contrastive models do not learn compositional representations: A simple Bag-of-Words (BoW) representation is completely sufficient to retrieve the right image from a batch, as the caption is typically long and detailed.
They have proposed solving this issue by training with hard negatives generated by following simple rules. Models trained on these negatives learn a compositional representation in the narrow domain defined by the set of rules used to generate hard negatives. 
Training with hard negatives has indeed been shown to yield in-distribution improvements, but these gains often fail to generalize to hard negatives that exhibit slightly different structures \citep{sugarcrepe,lewis2022does}.
At the same time, the improvements in compositionality achieved via hard negative training frequently come at the expense of generalization to other downstream tasks; precisely the property that originally made contrastive V\&L models popular.

Here, we follow a different approach and focus on better exploiting existing pre-training data. 
In particular, we focus on noun-phrases, which are groups of words that function as a concept in a sentence, e.g. ``a red couch" or ``the tall building", and define two auxiliary loss terms, a contrastive concept loss and a cross-attentive pooling loss, which combine to rectify the root causes of the contrastive learning behavior.
Our contrastive concept loss focuses on contrasting noun-phrases and images. Noun-phrases are short and thus are not solvable via a BoW representation. To provide an example: For the noun-phrase ``a red couch", a BoW representation is likely insufficient, as the probability of an image containing a couch and anything colored in red is quite high. Thus, the model must learn a latent representation that is more discriminative.
Since our method relies on real data positives instead of synthetically generated hard negatives, our method is less likely to learn shortcuts that do not generalize to broader domains.

Despite the shortcuts when training with long captions, the overall architectural design of contrastive V\&L models also hinders the learning of a good compositional representation.
Visual and textual representations are pooled into single-vector embeddings, which mix nouns and adjectives from different regions of the image or different parts of the caption. 
This global pooling operation results in a complete loss of associations between attributes and nouns.
Fine-tuning with post-pooling hard negative losses can yield improvements by aligning the model to in-domain distributions, however, this will not substantially change the underlying behavior of the encoders.

Instead
of focusing on encouraging binding after the global pooling with hard negatives, we focus on learning binding \textbf{before} global pooling.
Once learned, the model only needs to preserve a representation of that binding through the pooling operation.
To encourage binding before the pooling operation, we introduce a cross-modal attention pooling function that extracts noun-phrase-specific information from the full visual representation, thereby producing multi-word, concept-specific embeddings learned with an auxiliary contrastive loss.
As the attention-pooling function has no learnable parameters, this directly propagates the learning signal to the penultimate representation, i.e., it encourages a compositional representation before the global pooling operation. 
A schematic of our complete method can be seen in \cref{fig:method_overview}.

Our method, C$^{2}$LIP (Concept-centric CLIP), is simple to use, comes with no additional trainable parameters, and requires only fine-tuning for a few epochs without the incurred cost of complicated hard negative generation pipelines.
We show quantitatively that this simple method leads to improvements over its base model and reaches SOTA on the SugarCrepe \cite{sugarcrepe} and SugarCrepe++ \cite{sugarcrepe++} benchmarks.
At the same time, our method leads to improvements on image caption retrieval datasets while incurring only modest drops in classification performance on ImageNet. Note that the drop in performance on ImageNet can be attributed to two factors: 
First, fine-tuning is performed on a less diverse dataset than the one used for pre-training. 
Second, the shift toward a more scene centric representation (required for compositional reasoning) conflicts with ImageNet's objective of focusing on the most salient object.
Overall, C$^{2}$LIP demonstrates high performance across a variety of benchmarks and tasks. In contrast, the baselines usually excel at a single task but perform considerably poorer on the others.

In summary, our contributions are: 1) We introduce a hard negative-free pipeline to fine-tune existing contrastive V\&L models to improve compositionality. 2) We show that training on short noun-phrases instead of long detailed captions leads to improvements on compositionality while maintaining zero-shot capabilities. 3) We introduce a parameter-free cross-modal attention pooling mechanism that is used during training to encourage concept binding before the global pooling operation. 4) The proposed method is simple to train, adds minimal training overhead, and retains unchanged inference. 5) We show quantitatively that our method C\textsuperscript{2}LIP reaches SOTA on standard compositionality tasks, while at the same time maintaining high performance on standard retrieval and zero-shot benchmarks. It is particularly appealing that C\textsuperscript{2}LIP is among the top performers on all benchmarks and metrics tested, instead of excelling on some at the cost of others, as frequently observed for other methods.

\section{Related work}
\label{sec:realted_work}

Despite the great success of contrastive models for zero-shot classification and retrieval, several issues emerge when they are applied to more sophisticated tasks. 
For example, \citet{negclip_iclr23} showed that CLIP tends to learn BoW representations. Even more concerning, \citet{tang_emnlp23} found that CLIP matches an image of an eggplant and a lemon to the caption ``a purple lemon" rather than the correct ``a yellow lemon".

Various strategies have been tested to prevent contrastive models from learning BoW representations \citep{Singh2024-cn,ce-clip-Zhang2023-vj,Doveh2023-qp,Doveh2022-yg,Patel2024-qj,clic_Peleg2025}. Many of these follow \citet{negclip_iclr23} and use spaCy \cite{spacy2} to swap words in the captions, thereby creating hard negatives \citep{ce-clip-Zhang2023-vj,Doveh2023-qp,Doveh2022-yg, clic_Peleg2025} for the contrastive loss. Others achieve the same goal by employing a BERT model to replace words \citep{ce-clip-Zhang2023-vj,Doveh2022-yg}. 
More sophisticated methods generate hard negatives with an LLM, for instance via in-context learning \citep{Patel2024-qj}, however, this approach incurs the risk of failures during hard negative creation as well as a significant computational cost.
However, \citet{sugarcrepe,lewis2022does} showed that designing a caption augmentation pipeline that generalizes to unseen cases is difficult, and many approaches merely exploit biases in the test datasets. This questions whether the reliance on hand-crafted hard negatives is a promising direction in the first place. In parallel with hard negative captions, several works explore hard negative images generated by text-to-image models \citep{li_cvpr25,Patel2024-qj}.
These approaches typically generate hard negative captions and use these as input to a text-to-image model to obtain a hard negative image.
Producing a large number of hard‑negative images is very costly, as it requires both the creation of hard‑negative texts and the execution of the text‑to‑image model, and thus inflates the dataset size, leading to substantially higher computational demands.

Many of the methods described above rely on custom contrastive losses, such as an extension of the MIL-NCE loss \citep{Doveh2023-qp}, an adaptive margin loss \citep{li_cvpr25}, or a dynamic threshold loss \citep{ce-clip-Zhang2023-vj}. These losses are necessary to mitigate the negative impact of noisy or potentially erroneous synthetic samples, but each introduces its own set of challenges.
Other works \citep{dreamlip-zheng,Doveh2023-qp,gurung2025common,Liu2024-sv} argue that the problem lies in caption quality and therefore obtain dense captions \citep{dreamlip-zheng,Doveh2023-qp,Liu2024-sv} for the images. Unfortunately, generating such dense captions is expensive, and additional tricks \citep{dreamlip-zheng,Liu2024-sv} are required to make use of long captions in contrastive learning.

Another line of work attributes the problem to the model architecture, in particular the late interaction between modalities \citep{assouel2025object,xiao2025flair}. 
Inspired by object-centric learning, \citet{assouel2025object} use cross-attention to obtain caption-conditioned visual representations during both training and inference. 
However, their method requires an LLM to decompose a caption into a scene graph, effectively outsourcing most of the binding problem to the LLM, and requires an independent forward pass for each sub-caption both during training and inference, increasing the cost of training and inference drastically.
Our approach also uses attention to produce text-conditioned visual representations, but it is considerably simpler and does neither rely on an LLM, nor a complicated binding module, nor multiple forward passes.  
Instead, we only need to detect noun-phrases in regular captions (e.g., using spaCy \citep{spacy2}) to use in our two auxiliary training objectives.
The inference pipeline is identical to the vanilla CLIP/SigLIP pipeline.

\section{Concept-centric contrastive learning}

\subsection{Preliminaries: SigLIP model}
\label{ssec:siglip_prelim}

In this work, we utilize SigLIP \cite{siglip} as base model, as it already provides higher initial performance than CLIP. In this section we provide the necessary background on SigLIP by briefly describing the model architecture, especially its attention pooling layer and the loss function, which we exploit in our concept-aware objectives. However, our method does not require any component specific to SigLIP. Thus, it is in principle applicable to any CLIP-like model. We leave such explorations for future work.

The SigLIP model~\cite{siglip} follows the conventional contrastive learning framework with two neural networks: the vision encoder $f_{img}$ and text encoder $f_{txt}$ that project the input image-caption $(I, T)$ pair separately into two embedding vectors, $(v, t)$, in the joint subspace $\in {\mathbb{R}}^D$, where $D$ is the dimensionality of this subspace. The encoders are jointly trained by maximizing the cosine similarity of the positive pairs $s_{pos} = v_{pos} \cdot t_{pos}$, while at the same time minimizing the similarity scores $s_{neg}$ of negative pairs within a training batch. Particularly, the SigLIP model deviates from the preceding CLIP model~\cite{clip_openai} at two important points: visual attention pooling and sigmoid loss.
\\\\
\noindent\textbf{Visual attention pooling}. The CLIP vision encoder~\cite{clip_openai} adopted the standard vision transformer (ViT)~\cite{vit}, where a learnable \textit{[CLS]} token is appended to the input tokens and used to pool features via self-attention through all transformer layers. Whereas in SigLIP, a separate learnable token $q \in \mathbb{R}^D$ is employed to pool features \textit{after} the last transformer layer using an additional attention layer, as demonstrated in \cref{fig:siglip}. Let $V \in \mathbb{R}^{M \times D}$ be the outputs of the ViT, in which $M$ is the number of image patches. The visual embedding $v$ is calculated as follows:
\begin{align}
    \bar{q} &= f_{query}(q), \bar{K} = f_{key}(V), \bar{V} = f_{value}(V), \label{eq:v_proj}\\
    \tilde{q} &= \bar{V}^T \cdot \text{attn}(\bar{q}, \bar{K}), \label{eq:q_attn}\\
    v &= f_{MLP}(\tilde{q}) \label{eq:v_mlp},
\end{align}
where $f_{query}$, $f_{key}$, $f_{value}$ and $f_{MLP}$ are neural functions.
\\\\
\noindent\textbf{Contrastive sigmoid loss}. We consider a minibatch $B$ comprising paired embeddings $\{(v_i,t_i)|_{i=1}^{|B|}\}$, which we can organize into $|B|$ positive and $|B| \times (|B| - 1)$ negative pairs in total. We define an indicator function $z_{ij}$ with value 1 if $i=j$ and $-1$ otherwise, reflecting the assumption of pairwise matches between images and captions. The contrastive sigmoid loss is given by
\begin{equation}
\mathcal{L}_{\text{contrastive}} = - \frac{1}{|B|} \sum_{i=1}^{|B|} \sum_{j=1}^{|B|} \text{log}\ \sigma \left( z_{ij} ( \tau {v_i} \cdot {t_j} + b ) \right),
\label{eq:contrastive}
\end{equation}
where $\sigma$ denotes the sigmoid function, $\tau$ and $b$ are learnable scalars corresponding to scale and bias terms. 
The sigmoid loss improves computational efficiency when compared to the more costly batch-wise softmax calculation required by CLIP.
However, training only with this objective yields a BoW representation.
In the next section we introduce additional concept-aware contrastive losses and explain how these losses are combined with our noun-phrase captions and cross-attention pooling function to promote learning a compositional representation.
Note that these modifications do not require any changes to the original model architecture and do not alter the inference procedure.

\subsection{Concept-aware contrastive training}
\label{ssec:concept_aware_training}

We emphasize the binding of the elements in \textit{noun-phrase concepts}, and aligning such concepts to the corresponding images. We start by identifying a set of noun-phrases from each input caption, which can be done offline using standard NLP tools, e.g. spaCy \citep{spacy2}.
\\\\
\noindent\textbf{Concept-aware contrastive loss.} The objective of this loss is to explicitly encourage the encoders to bind elements of a noun-phrase together. More specifically, each caption $T_i$ contains $K_i$ concepts (noun-phrases), with latent representations $\{\textbf{c}_{i}^k |_{k=1}^{K_i} \}$. 
These representations are computed by pooling the representations of the text tokens corresponding to the respective noun-phrase, as depicted in \cref{fig:our_method}.
We then devise an image-text contrastive loss similar to \cref{eq:contrastive}. However, different from \cref{eq:contrastive}, we match each image to all of its corresponding sub-captions (i.e. noun-phrase concepts) simultaneously. 
This enforces that the model represents all concepts in the global image representation.
Our loss extends the standard SigLIP loss to a version that accepts multiple positives per sample, thus the indicator function $z_{ij}^\prime$ must be prepared accordingly.
The \textbf{n}oun-\textbf{p}hrase \textbf{c}oncept (npc) loss is given by
\begin{equation}
\mathcal{L}_{npc} = - \frac{1}{\mathcal{K}} \sum_{i=1}^{|B|} \sum_{j=1}^{\mathcal{K}} \text{log}\ \sigma \left( {z}_{ij}^\prime ( \tau {v_i} \cdot {c_j} + b ) \right),
\label{eq:l_npc}
\end{equation}

\noindent where $\mathcal{K} = \sum_{i=1}^{|B|} K_i$.
\\\\
\noindent\textbf{Cross-attended concept-aware loss.}
In addition to SigLIPs attention-pooling in the vision tower, we apply cross-modal attention-pooling to extract information from the visual tokens using the concept embeddings $c$ as queries, as shown in \cref{fig:our_method}. 
We reuse the projected value $\bar{V}$ in \cref{eq:v_proj} and the $MLP$ in \cref{eq:v_mlp} to project the visual tokens to
$\bar{V}^{\prime} = f_{MLP}(\bar{V})$ in the joint subspace, which are used as key and value in the cross-attention function:
\begin{equation}
    \hat{v}(c) = {{\bar{V}^{\prime T}}}  \cdot \text{attn}(c, \bar{V}^{\prime})
    \label{eq:crossmodal_pool}.
\end{equation}
Note that this attention-pooling does not add any additional parameters, and it is only utilized during training. 

The \textbf{cross}-\textbf{a}ttended \textbf{c}oncept-aware ($xac$) loss is defined similarly to $\mathcal{L}_{npc}$ in \cref{eq:l_npc}, but replacing the unimodal visual latent representation $v$ with the cross-modal concept-attended variant $\hat{v}(c)$, as follows:
\begin{equation}
\mathcal{L}_{xac} = - \frac{1}{\mathcal{K}} \sum_{i=1}^{|B|} \sum_{j=1}^{\mathcal{K}} \text{log}\ \sigma \left( {z}_{ij}^\prime ( \tau {\hat{v}_i(c_j)} \cdot {c_j} + b ) \right).
\label{eq:L_xac}
\end{equation}

\noindent\textbf{Total training loss. }
Finally, the model is fine-tuned with the combined loss given by
\begin{equation}
    \mathcal{L}_{total} = \mathcal{L}_{contrastive} + \lambda_{npc} \mathcal{L}_{npc} + \lambda_{xac} \mathcal{L}_{xac},
    \label{eq:total}    
\end{equation}
where  $\lambda_{\text{npc}}$ and $\lambda_{\text{xac}}$ are trade-off hyper-parameters.

\section{Experimental results}
We detail our experimental settings, baselines, datasets and the evaluation protocols in \cref{ssec:experimental_setting}. We present and discuss our quantitative results in \cref{ssec:quantitative_evaluation}, provide qualitative results in \cref{ssec:qualitative_results} and ablate the components of our method in \cref{ssec:ablations}.

\subsection{Experimental setting}\label{ssec:experimental_setting}

We fine-tune the pretrained SigLIP on CC3M \citep{sharma2018conceptual} using our proposed method.
Particularly, we use the new short captions of CC3M introduced in DreamLIP~\cite{dreamlip-zheng}, which contain richer descriptions of objects and attribute bindings. 
We fine-tune the model for five epochs using Adam optimizer with base learning rate of 1e-5, on eight A40 GPUs with effective batch size of 768. The hyper-parameters $\lambda_{npc}$ and $\lambda_{xac}$ in \cref{eq:total} are empirically chosen as 1 and 0.01, respectively. Our implementation extends from \textit{OpenCLIP} \citep{ilharco_gabriel_2021_5143773}.
\par
\vspace{\baselineskip}
\noindent\textbf{Data pre-processing.} First, we use spaCy for dependency parsing of the original captions, and traverse the dependency trees to extract the noun-phrase concepts.
\par
\vspace{\baselineskip}
\noindent\textbf{Baselines.} We use a ViT-B/16 in all our experiments. Accordingly, we compare with baselines utilizing the same model, or models with similar parameter count.  
We select checkpoints of corresponding models trained/fine-tuned on CC3M whenever available, and use any provided checkpoints otherwise.
Refer to \cref{tab:training_data} for details on the training setup of our baselines.

Our main baseline is the SigLIP model pretrained on WebLI and fine-tuned on the DreamLIP variant of CC3M. 
This results in a fair baseline that has been trained with exactly the same data, exactly the same training setup and allows a comparison without confounding factors like the effect of a narrower domain of CC3M in comparison to WebLI.
Despite that, we also compare performance of our model to that of larger models using a ViT-L encoder, as well as models trained with multi-task objectives. 
\par
\vspace{\baselineskip}
\noindent \textbf{Evaluation protocol.} To ensure fair, consistent performance comparisons and facilitate reproducibility, we evaluate all methods with the \textit{CLIP\_Benchmark} tool \citep{cherti_2025_15403103} in the same environment, and report the resulting metrics on a variety of benchmarks:
\begin{itemize}
    \item \textit{Compositionality}: SugarCrepe and SugarCrepe++
    \item \textit{Zero-shot classification}: ImageNet1K
    \item \textit{Zero-shot retrieval}: Flickr30K and MSCOCO
    \item \textit{Fine-grained retrieval}: DOCCI~\cite{OnoeDocci2024} and ImageInWords (IIW)~\cite{garg2024imageinwords}. Following the experiments of \citet{xiao2025flair}, we split the long captions into single sentences and evaluate retrieval performance on them.
\end{itemize}

\vspace{\baselineskip}
\noindent\textbf{Evaluation metrics.}
For evaluation of compositionality, we follow the standard evaluation protocol of SugarCrepe and SugarCrepe++ and report accuracy. These benchmarks consider a sample as correct, if the true caption is assigned a higher similarity to the image than the hard negative caption. 
The datasets allow to evaluate for different word types separately, namely for \textit{objects}, \textit{attributes} and \textit{relations of concepts}. 
Accuracy is always reported for a set of subtasks, differing in how the false samples are created: ``Add'' creates the false caption by adding an unrelated word of a given word type, ``Swap'' swaps a given word type with the same word type within the caption and ``Replace'' replaces a word with a new word of the same type. All evaluations on SugarCrepe are image-to-text (I2T).

SugarCrepe++ extends the protocol of SugarCrepe by using two positive captions and requiring both of them to be higher ranked than the negative, to be considered correct.
Furthermore,
SugarCrepe++ adds the text-to-text tasks (TOT), which probes the text encoders in isolation for compositionality, following the same evaluation pattern.

For the remaining tasks we follow the default evaluation settings of \textit{CLIP\_Benchmark} \citep{cherti_2025_15403103}:
For ImageNet we report accuracy. For Flickr30k, MSCOCO, DOCCI and IIW we report Recall@5.

\begin{table}[ht]
\centering
\caption{\textbf{Training data of baselines.} Summary of baseline methods and their corresponding training datasets, excluding the original CLIP and SigLIP models. The last column indicates whether the provided checkpoints were trained from scratch (\checkmark).}
\label{tab:training_data}
\resizebox{\columnwidth}{!}{
\begin{tabular}{l|c|c}
\toprule
\multicolumn{1}{c|}{Model}               & Training data          & Train from scratch \\\midrule
\textit{\textbf{Composition-aware}}     & \multicolumn{1}{l|}{}   &                    \\
\hspace{1em}CE-CLIP~\cite{ce-clip-Zhang2023-vj}                                 & MSCOCO                   &                    \\
\hspace{1em}NegCLIP~\cite{negclip_iclr23}                                 & MSCOCO                   &                    \\
\hspace{1em}CLIC~\cite{clic_Peleg2025}                                    & LAION-1.5B             &                    \\\cmidrule{2-2}
\hspace{1em}DAC~\cite{Doveh2023-qp}                                     & \multirow{10}{*}{CC3M} &                    \\
\hspace{1em}SLVC~\cite{Doveh2022-yg}                                    &                        &                    \\
\hspace{1em}CoN-CLIP~\cite{Singh2024-cn}                                &                        &                    \\
\hspace{1em}TripletCLIP~\cite{Patel2024-qj}                             &                        & \checkmark                  \\
\textit{\textbf{Codebook-based}}        &                        &                    \\
\hspace{1em}Codebook-CLIP~\cite{codebook-clip-chen2023}                           &                        & \checkmark                  \\
\hspace{1em}IL-CLIP~\cite{il-clip-zheng2024iterated}                                 &                        & \checkmark                  \\
\textit{\textbf{Fine-grained training}} &                        &                    \\
\hspace{1em}DreamLIP-3m~\cite{dreamlip-zheng}                             &                        & \checkmark                  \\
\hspace{1em}FLAIR-3m~\cite{xiao2025flair}                                &                        & \checkmark                  \\\cmidrule{2-2}
\hspace{1em}FG-CLIP~\cite{fg-clip-Xie2025}                                 & LAION-2B + FineHARD    & \checkmark                  \\
\hspace{1em}FineCLIP~\cite{fineclip-dong}                                & MSCOCO                   &                    \\
\hspace{1em}LLIP~\cite{llip}                                            & Common Crawl 12.8B        & \checkmark        \\
\textit{\textbf{Large-sized models}}    &                        &                    \\
\hspace{1em}CLIP-A (ViT-L/14)~\cite{Li2023-kx}                       & LAION-400M             & \checkmark                  \\
\hspace{1em}CLIPS (ViT-L/14)~\cite{Liu2024-sv}                        & Recap-DataComp-1B     & \checkmark                 \\
\midrule
\textit{\textbf{Multi-task training}}    &                        &                    \\
\hspace{1em}BLIP-B~\cite{li2022blip}     &      14M samples                   &     -               \\
\hspace{1em}FLAVA~\cite{singh2022flava}     &   PMD-80M                      &      -              \\
\hspace{1em}SigLIP-2~\cite{siglip2}     &     WebLI                    &           -        \\
\bottomrule
\end{tabular}
}
\end{table}

\subsection{Quantitative evaluation}
\label{ssec:quantitative_evaluation}

\begin{table*}[ht]
\centering
\caption{\textbf{Performance comparison of \textit{ViT-B-based} models}. C\textsuperscript{2}LIP achieves consistently high performance on SugarCrepe and SugarCrepe++ compared to the composition-aware models, despite only relying on regular captions during training. Additionally, our model is able to maintain and improve the retrieval and zero-shot capabilities of the original SigLIP, achieving the best overall score.}
\label{tab:retrieval_bench}
\resizebox{\textwidth}{!}{
\begin{tabular}{l|ccc|cccc|c|c|c|c|c|c}
\toprule
\multicolumn{1}{c|}{\multirow{3}{*}{Models}} & \multicolumn{3}{c|}{SugarCrepe}                                          & \multicolumn{4}{c|}{SugarCrepe++}                       & \multirow{3}{*}{ImNet1K} & \multirow{3}{*}{Flickr30k} & \multirow{3}{*}{MSCOCO} & \multirow{3}{*}{DOCCI} & \multirow{3}{*}{IIW} & \multirow{3}{*}{\textbf{Average}} \\
\cmidrule(l{1pt}r{1pt}){2-4}\cmidrule(l{1pt}r{1pt}){5-8}
\multicolumn{1}{c|}{}                        & \multirow{2}{*}{Add} & \multirow{2}{*}{Replace} & \multirow{2}{*}{Swap} & \multicolumn{2}{c}{Replace} & \multicolumn{2}{c|}{Swap} &         &       &      &       &                      \\
\cmidrule(l{1pt}r{1pt}){5-6}\cmidrule(l{1pt}r{1pt}){7-8}
\multicolumn{1}{c|}{}                        &                      &                          &                       & I2T          & TOT          & I2T         & TOT        &                             &                             &                       &                        &                      &      \\
\midrule
SigLIP ViT-B/16                             & 86.5                 & 84.1                     & 65.8                  & 73.8         & 62.8         & 48.0        & 34.7       & \textbf{76.1}                        & 95.2                        & 78.9                  & 58.9                   & 75.3                 & 70.0                         \\
CLIP (OpenAI) ViT-B/32                      & 73.0                 & 80.0                     & 62.7                  & 69.5         & 60.5         & 45.7        & 27.4       & 63.3                        & 89.0                        & 65.4                  & 47.1                   & 69.5                 & 62.8                         \\
CLIP (OpenAI) ViT-B/16                      & 72.7                 & 80.4                     & 62.5                  & 70.1         & 60.2         & 43.8        & 23.9       & 68.4                        & 90.9                        & 67.6                  & 50.4                   & 71.3                 & 63.5                         \\

\midrule
\textbf{\textit{Fine-grained models}} & & & & & & & & & & & & & \\
FG-CLIP                                     & 84.7                 & 85.1                     & 69.9                  & 75.8         & 67.5         & 51.5        & 38.2       & 69.0                        & \underline{95.8}                        & 78.4                  & 56.7                   & 75.6                 & 70.7                         \\
FineCLIP                                    & 85.4                 & 85.3                     & 66.8                  & 72.8         & 68.7         & 43.9        & 26.8       & 55.8                        & 93.5                        & 79.0                  & 44.9                   & 67.1                 & 65.8                         \\
DreamLIP-3m                                 & 72.9                 & 77.5                     & 64.2                  & 61.2         & 51.5         & 44.4        & 30.1       & 31.6                        & 83.1                        & 61.2                  & 58.8                   & \textbf{77.9}                 & 59.5                         \\
FLAIR-3m                                    & 80.6                 & 81.4                     & 70.9                  & 66.3         & 57.5         & 49.5        & 35.2       & 33.7                        & 90.4                        & 71.5                  & 47.2                   & 72.8                 & 63.1                         \\
LLIP                                        & 71.4                 & 78.9                     & 57.8                  & 67.3         & 56.8         & 40.8        & 27.3       & 60.8                       & 90.2                      & 67.5                  & 46.9                   & 70.7                 & 61.4      \\
\midrule
\textbf{\textit{Codebook-based models}} & & & & & & & & & & & & & \\
Codebook-CLIP                               & 50.6                 & 54.5                     & 47.5                  & 34.3         & 13.8         & 28.3        & 10.1       & 0.0                         & 0.6                         & 0.1                   & 0.1                    & 0.7                  & 20.0                         \\
IL-CLIP                                     & 51.8                 & 52.8                     & 55.2                  & 34.4         & 31.8         & 36.9        & 14.7       & 0.0                         & 0.4                         & 0.1                   & 0.1                    & 0.7                  & 23.2                         \\
\midrule
\textbf{\textit{Composition-aware models}} & & & & & & & & & & & & & \\
CE-CLIP                                     & 92.9                 & 87.0                     & 74.9                  & 56.5         & 67.0         & 34.3        & 32.5       & 40.4                        & 87.4                        & 71.9                  & 30.8                   & 50.1                 & 60.5                         \\
NegCLIP                                     & 85.8                 & 85.0                     & \textbf{75.3}         & 69.1         & \underline{70.9}         & 53.4        & \underline{39.1}       & 55.7                        & 92.4                        & 73.9                  & 45.2                   & 66.4                 & 67.7                         \\
CLIC                                      & 89.8                 & 85.8                     & 72.5                  & \underline{76.6}  & 58.1         & \textbf{59.0} & 27.0       & 66.6                        & 91.1                        & 67.4                  & 51.3                   & 73.4                 & 68.2                         \\
DAC-SAM                                     & 91.6                 & 87.0                     & 73.5                  & 52.4         & 60.2         & 30.6        & 18.4       & 52.3                        & 84.0                        & 58.8                  & 31.2                   & 50.6                 & 57.5                         \\
DAC-LLM                                     & \underline{93.7}        & \textbf{89.5}            & \underline{74.6}                  & 53.7         & 59.6         & 32.2        & 18.1       & 51.1                        & 83.7                        & 59.0                  & 27.1                   & 43.9                 & 57.2                         \\
SLVC-R                                      & 85.4                 & 78.9                     & 68.8                  & 64.0         & 68.2         & 51.3        & 28.4       & 58.5                        & 90.1                        & 66.9                  & 42.1                   & 66.1                 & 64.1                         \\
SLVC-RL                                     & 78.5                 & 75.9                     & 65.5                  & 61.9         & 70.0         & 46.0        & 26.3       & 59.7                        & 90.0                        & 67.0                  & 41.8                   & 67.3                 & 62.5                         \\
CoN-CLIP                                    & 82.4                 & 77.4                     & 62.4                  & 68.1         & 70.1         & 45.2        & 28.0       & 63.7                        & 86.0                        & 61.2                  & 46.0                   & 67.1                 & 63.1                         \\
TripletCLIP                                 & 88.3                 & 87.0                     & 69.9                  & 69.9         & 69.4         & 41.4        & 28.5       & 45.9                        & 81.7                        & 54.4                  & 39.8                   & 62.5                 & 61.6                         \\
\midrule
\textbf{\textit{Models fine-tuned by us}} & & & & & & & & & & & & & \\
SigLIP ViT-B/16 (ft. CC3M)                  & 87.9                 & 85.6                     & 69.7                  & 73.5         & 67.9         & 49.8        & 36.6       & \underline{75.9}                        & 95.6                        & \underline{80.3}                  & \underline{59.8}                   & 75.5                 & \underline{71.5}                         \\
\textbf{C\textsuperscript{2}LIP}           & \textbf{94.2}                 & \underline{88.3}                     & 73.1                  & \textbf{79.7}         & \textbf{75.3}         & \underline{55.2}        & \textbf{44.2}       & 73.5                        & \textbf{97.0}                        & \textbf{82.7}                  & \textbf{60.0}                   & \underline{76.4}                 & \textbf{75.0}                         \\
\bottomrule
\end{tabular}
}
\end{table*}

\noindent\textbf{Comparison with ViT-B baselines}.
Our main results are shown in \cref{tab:retrieval_bench}.
The last two rows show the most important results as they show the most comparable models, both trained by us under the exact same setting on the DreamLIP variant of CC3M with either the standard SigLIP objective or our objectives.
It can be seen that fine-tuning on DreamLIP-CC3M alone leads only to marginal improvements with respect to the original SigLIP checkpoint. 
However, when using our method we observe considerable improvements on all compositionality benchmarks and tasks. Additionally, our model improves on  Flickr30k, MSCOCO, and IIW. Thus, it improves both long and short caption retrieval.
Only on ImageNet does the performance drop. This phenomenon is observed for most composition-aware models. Compared to these methods, the drop on ImageNet for C$^2$LIP is very small.
We attribute the drop in performance to the more scene centric representation encouraged by our training method, which can lead to poorer results on tasks like ImageNet classification, an extremely object centric task, for which it is beneficial to ignore everything except for the most salient and central object. 

Compared to previous methods that aimed at improving compositionality (composition-aware models in \cref{tab:retrieval_bench}), C$^2$LIP shows consistently high performance on all benchmarks, making it the new SOTA method among models with similar backbone capacity.
While C$^2$LIP is outperformed on some metrics by either NegCLIP or DAC‑LLM on the SugarCrepe benchmark (which is known to be hackable), it outperforms both methods on the more trustworthy SugarCrepe++ benchmark. Similarly, CLIC surpasses C$^2$LIP on SugarCrepe++ Swap-I2T task, but a closer look at text-only tasks (TOT) reveals that CLIC performs poorly, exposing major limitations in the compositional representation learned by the CLIC text encoder.

In summary, C$^2$LIP demonstrates overall SOTA performance, achieving the best or second best results on most benchmarks and metrics while remaining among the top models on the remaining ones. C$^2$LIP stands out particularly for its consistently high performance across all benchmarks, whereas other models tend to excel on a single benchmark but perform considerably poorer on others.  With this reliable performance, C$^2$LIP is an excellent post-training method to improve compositional comprehension in foundation models, without sacrificing performance on other tasks.

\begin{table*}[ht]
\centering
\caption{\textbf{Attribute-binding performance comparison.} We compare the accuracy of C\textsuperscript{2}LIP with composition-aware baselines on \textbf{\textit{attribute}} replacement and swapping experiments of SugarCrepe and SugarCrepe++. Our model consistently outperforms the competing baselines on most tasks and attains the best overall score. Moreover, previous methods rely on hard negatives to induce a compositional representation, which often harms generalization to natural data and leads to a considerable performance drop on SugarCrepe++. In contrast, C\textsuperscript{2}LIP maintains high performance on all metrics, further highlighting the strength of our proposed approach.}
\label{tab:attribute}
\small
\begin{tabular}{l|cc|cccc|c}
\toprule
\multicolumn{1}{c|}{\multirow{2}{*}{Model}} & \multicolumn{2}{c|}{SugarCrepe} & \multicolumn{4}{c|}{SugarCrepe++}                              & \multirow{2}{*}{\textbf{Average}} \\
\cmidrule(l{1pt}r{1pt}){2-3}\cmidrule(l{1pt}r{1pt}){4-7}
\multicolumn{1}{c|}{}                       & Replace        & Swap          & Replace-I2T   & Replace-TOT   & Swap-I2T      & Swap-TOT      &                          \\
\midrule
SigLIP ViT-B/16                            & 86.7           & 71.5          & 75.5          & 64.2          & 56.3          & 46.4          & 66.8                     \\
\midrule
CE-CLIP                                     & 88.8           & {\ul 77.0}    & 52.0          & 64.2          & 32.3          & 36.3          & 58.4                     \\
NegCLIP                                     & 85.9           & 75.4          & 67.1          & 70.7          & 55.0          & {\ul 50.3}    & {\ul 67.4}               \\
CLIC                                       & 86.6           & 74.0          & {\ul 75.9}    & 52.4          & {\ul 62.2} & 31.2          & 63.7                     \\
DAC-SAM                                    & 85.9           & 75.1          & 44.3          & 56.0          & 33.5          & 25.4          & 53.3                     \\
DAC-LLM                                    & \textbf{89.5}  & 74.2          & 47.7          & 59.5          & 32.9          & 24.8          & 54.8                     \\
SLVC-R                                     & 81.4           & 69.1          & 61.6          & 67.4          & 53.2          & 36.3          & 61.5                     \\
SLVC-RL                                    & 76.8           & 66.5          & 57.1          & 67.0          & 48.8          & 34.3          & 58.4                     \\
CoN\_CLIP                                  & 79.7           & 66.1          & 68.1          & 64.7          & 50.3          & 37.2          & 61.0                     \\
TripletCLIP                                & 85.7           & 69.7          & 66.0          & {\ul 71.1}    & 44.4          & 38.1          & 62.5                     \\
\midrule
\textbf{C\textsuperscript{2}LIP}                                     & {\ul 89.2}     & \textbf{78.8} & \textbf{79.8} & \textbf{80.2} & {\bf 66.2}    & \textbf{59.8} & \textbf{75.7}   \\
\bottomrule        
\end{tabular}
\end{table*}

\begin{table*}[h]
\centering
\caption{\textbf{Comparison with SOTA foundation models.} We compare performance of C\textsuperscript{2}LIP with larger contrastive models (ViT-L), and SOTA models trained with multimodal/multi-task objectives (BLIP, FLAVA, SigLIP-2). Despite the substantial disadvantages in model size and training, C\textsuperscript{2}LIP has the best performance on ``Add'' and ``Replace'' tasks of SugarCrepe, and maintains close performance on other benchmarks. On average our model is only $1.7$ percentage points below the best model, CLIPS, which has 5x more parameters.}
\label{tab:compare_sota}
\resizebox{\textwidth}{!}{
\begin{tabular}{l|ccc|cccc|c|c|c|c|c|c}
\toprule
\multicolumn{1}{c|}{\multirow{3}{*}{Models}} & \multicolumn{3}{c|}{SugarCrepe}                                          & \multicolumn{4}{c|}{SugarCrepe++}                       & \multirow{3}{*}{ImNet1K} & \multirow{3}{*}{Flickr30k} & \multirow{3}{*}{MSCOCO} & \multirow{3}{*}{DOCCI} & \multirow{3}{*}{IIW} & \multirow{3}{*}{\textbf{Average}} \\
\cmidrule(l{1pt}r{1pt}){2-4}\cmidrule(l{1pt}r{1pt}){5-8}
\multicolumn{1}{c|}{}                        & \multirow{2}{*}{Add} & \multirow{2}{*}{Replace} & \multirow{2}{*}{Swap} & \multicolumn{2}{c}{Replace} & \multicolumn{2}{c|}{Swap} &                             &                             &                       &                     &         \\
\cmidrule(l{1pt}r{1pt}){5-6}\cmidrule(l{1pt}r{1pt}){7-8}
\multicolumn{1}{c|}{}                        &                      &                          &                       & I2T          & TOT          & I2T         & TOT        &                             &                             &                       &                        &                      &      \\
\midrule
\textbf{\textit{Multi-task SOTA}} & & & & & & & & & & & & & \\
BLIP-B                  & \underline{92.2} & 84.1 & 73.6 & 76.7 & \textbf{86.0} & 52.1 & 41.3 & 49.2 & \underline{97.2} & 79.1 & 51.6 & 72.4 & 71.3 \\
FLAVA                   & 79.6 & 84.9 & \underline{73.8} & 74.3 & 75.0 & \underline{56.6} & \textbf{51.0} & 56.9 & 91.8 & 73.9 & 48.3 & 72.9 & 69.9 \\
SigLIP2-ViT-B/16                            & 89.0                 & 86.0                     & 71.9                  & \underline{77.7}         & 69.7         & \textbf{57.9}        & 42.0       & 78.5                        & 96.5                        & \textbf{82.8}                  & 62.2                   & \underline{78.1}                 & 74.4                         \\

\midrule
\textbf{\textit{Larger models}} & & & & & & & & & & & & & \\
CLIP-A (ViT-L/14)                           & 85.8                 & 82.9                     & 63.8                  & 74.7         & 74.8 & 45.8        & 30.8       & \underline{79.6}                        & 95.2                        & 78.1                  & 59.4                   & 75.4                 & 70.5                         \\
CLIPS (ViT-L/14)                            & 86.6                 & \underline{86.1}                     & \textbf{77.8}                  & 76.6 & 66.6         & 59.4        & \underline{44.3}       & 76.8                        & \textbf{97.7}                        & 82.1                  & \textbf{66.2}                   & \textbf{81.8}                 & \textbf{75.2}                         \\
SigLIP ViT-L/16-res256                      & 86.7                 & 84.0                     & 72.8                  & 75.4         & 64.9         & 55.6        & 42.1       & \textbf{80.5}                        & 96.7                        & 82.3                  & \underline{62.5}                   & 77.6                 & 73.4                         \\

\midrule

\textbf{C\textsuperscript{2}LIP}           & \textbf{94.2}                 & \underline{88.3}                     & 73.1                  & \textbf{79.7}         & \underline{75.3}         & 55.2        & 44.2       & 73.5                        & 97.0                        & \underline{82.7}                  & 60.0                   & 76.4                 & \underline{75.0}                         \\
\bottomrule
\end{tabular}
}
\end{table*}

\vspace{\baselineskip}
\noindent\textbf{Comparison on attribute-binding performance.}
As shown in the previous section, C\textsuperscript{2}LIP improves all compositionality related tasks; however, the main gain lies in strengthening attribute-object binding.
Therefore, we examine the results on attribute–object binding in more detail.
As can be seen in \cref{tab:attribute}, on average, C\textsuperscript{2}LIP outperforms all prior composition-aware methods for attribute related tasks.
C\textsuperscript{2}LIP achieves the best results on almost every metric, being surpassed only by DAC-LLM on the SugarCrepe ``Replace'' task by $0.3$ percentage points.
Overall, C\textsuperscript{2}LIP shows consistently high performance, while baselines perform poorly on at least one of the tasks.

\par

\par
\vspace{\baselineskip}
\noindent\textbf{Comparison with models that use captioning and/or an LLM component.}
Contrastive learning offers clear advantages over models trained with captioning losses, primarily due to its inexpensive zero-shot capabilities and the strictly separate image and text encoders; properties that, i.e., BLIP \citep{li2022blip} does not possess. 
Nevertheless, captioning based methods remain popular, and a comparison is instructive. 
Similarly, LLM based approaches such as FLAVA \citep{singh2022flava} are rapidly gaining traction. 
We compare C\textsuperscript{2}LIP with models that incorporate a captioning loss and with FLAVA in the top part of \cref{tab:compare_sota}. 
It can be seen that C\textsuperscript{2}LIP is competitive even against these methods.

\par
\vspace{\baselineskip}
\noindent\textbf{Comparison with larger models.}
Next, we compare against contrastively trained models that have larger parameter count. The results are shown in the middle part of \cref{tab:compare_sota}. 
Again, C\textsuperscript{2}LIP performs competitively on compositionality tasks despite being considerably smaller.

\subsection{Qualitative Results}\label{ssec:qualitative_results}

\begin{figure}
    \centering
     \begin{subfigure}[b]{0.49\textwidth}
        \includegraphics[width=0.49\linewidth,trim={1.75cm 0.5cm 1.75cm 0},clip]{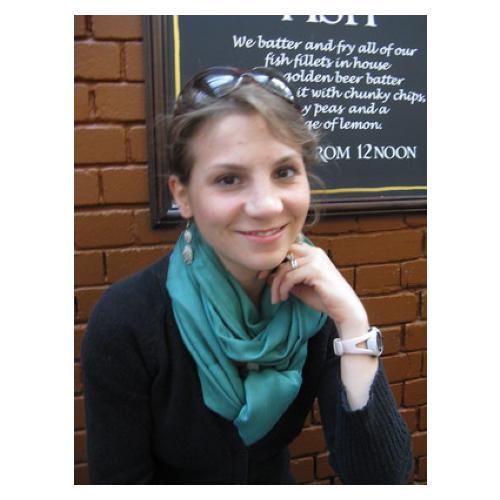}
        \includegraphics[width=0.49\linewidth,trim={1.75cm 0.5cm 1.75cm 0},clip]{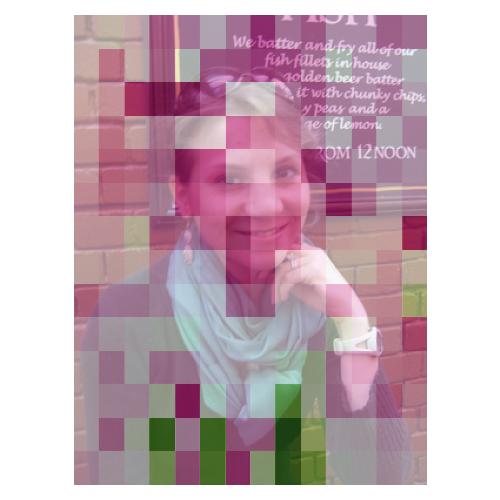}
        \caption{\textbf{Caption:} ``Black sweater''}
        \label{subfig:black_sweater}
    \end{subfigure}
    \begin{subfigure}[b]{0.49\textwidth}
        \centering
        \includegraphics[width=0.49\linewidth,trim={0cm 2.25cm 0cm 2.5cm},clip]{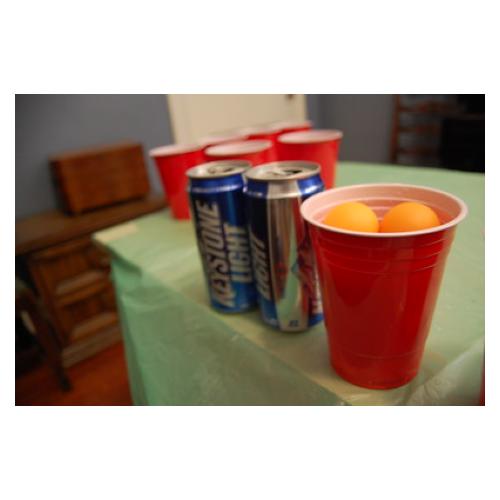}
        \includegraphics[width=0.49\linewidth,trim={0cm 2.25cm 0cm 2.5cm},clip]{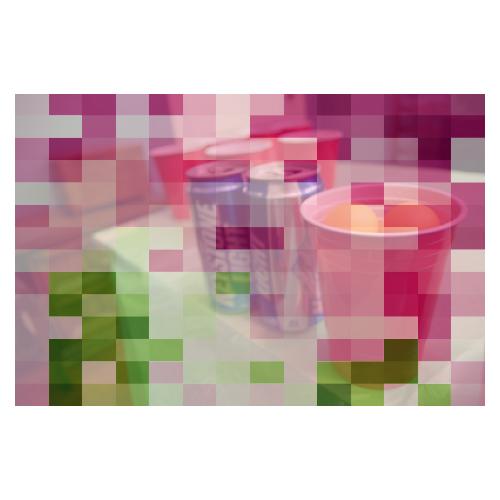}
        \caption{\textbf{Caption:} ``Green tablecloth''}
        \label{subfig:tablecloth}
    \end{subfigure}
    
    \caption{\textbf{Change in attention when using  C\textsuperscript{2}LIP} compared to SigLIP. We visualize the {\ul \textit{difference in attention}} between C\textsuperscript{2}LIP and SigLIP to the visual tokens given a caption and an image. Higher attention for C\textsuperscript{2}LIP is shown as \textcolor{OliveGreen}{green}. Lower  attention for C\textsuperscript{2}LIP indicated with \textcolor{RedViolet}{violet}. White means no change. \textbf{(a)} Black regions that are not a sweater get reduced attention, the sweater gets more or unchanged attention. \textbf{(b)} The background, the cups and beer cans get attended less after training with our method, while attention to the green tablecloth increases or stays the same.}
    \label{fig:attention_change}
\end{figure}

\begin{table*}[h]
\centering
\caption{\textbf{Training objective ablation.} Improvements in \textcolor{OliveGreen}{(green)} with respect to the variant fine-tuned on CC3M. \textbf{\textit{npc}}: noun-phrase loss, \textbf{\textit{xac}}: cross-attended concept-aware loss. Both components lead to significant gains.}
\label{tab:ablation}
\resizebox{\textwidth}{!}{
\begin{tabular}{l|ccc|cc|cc|c|c}
\toprule
\multicolumn{1}{c|}{\multirow{3}{*}{Training Configurations}}         & \multicolumn{8}{|c|}{SugarCrepe}                                                                                                                                                                                                                    & \multicolumn{1}{|c}{\multirow{3}{*}{Flickr8K}} \\
\cmidrule{2-9}
                                            & \multicolumn{3}{|c|}{Replace}                                                               & \multicolumn{2}{|c|}{Swap}                                   & \multicolumn{2}{|c|}{Add}                                    &   \multicolumn{1}{|c|}{\multirow{2}{*}{\textbf{Average}}}                          &  \multicolumn{1}{|c}{}            \\
\cmidrule(l{1pt}r{1pt}){2-4}\cmidrule(l{1pt}r{1pt}){5-6}\cmidrule(l{1pt}r{1pt}){7-8}
\multicolumn{1}{c|}{} & \multicolumn{1}{c}{Object} & \multicolumn{1}{c}{Attribute} & \multicolumn{1}{c|}{Relation} & \multicolumn{1}{c}{Object} & \multicolumn{1}{c|}{Attribute} & \multicolumn{1}{c}{Object} & \multicolumn{1}{c}{Attribute} & \multicolumn{1}{|c|}{} &  \multicolumn{1}{|c}{} 
\\\midrule
Original SigLIP ViT-B/16                    & 95.3                       & 86.7                          & 70.3                         & 60.0                       & 71.5                          & 89.1                       & 83.8                          & 79.5                        & 94.9         \\
\hspace{1em}+ fine-tuned on CC3M                         & 95.8                       & 87.4                          & 73.5                         & 65.3                       & 74.2                          & 88.9                       & 87.0                          & 81.7                        & 95.2                     \\
\hspace{2em}+ npc                                        & \textbf{96.7} \textcolor{OliveGreen}{(+0.9)}                     & \textbf{89.7} \textcolor{OliveGreen}{(+2.3)}                   & 78.5 \textcolor{OliveGreen}{(+5.0)}                        & 66.5 \textcolor{OliveGreen}{(+1.2)}                      & 78.5 \textcolor{OliveGreen}{(+4.3)}                         & 93.0 \textcolor{OliveGreen}{(+4.1)}                & 93.9 \textcolor{OliveGreen}{(+6.9)}                          & 85.3 \textcolor{OliveGreen}{(+3.6)}                       & 96.3 \textcolor{OliveGreen}{(+1.1)}               \\
\hspace{3em}+ xac (\textit{full})                                       & \textbf{96.7} \textcolor{OliveGreen}{(+0.9)}                      & 89.2 \textcolor{OliveGreen}{(+1.8)}                          & \textbf{78.9} \textcolor{OliveGreen}{(+5.4)}                   & \textbf{67.3} \textcolor{OliveGreen}{(+2.0)}                      & \textbf{78.8} \textcolor{OliveGreen}{(+4.6)}                         & \textbf{93.5} \textcolor{OliveGreen}{(+4.6)}             & \textbf{94.9} \textcolor{OliveGreen}{(+7.9)}                 & \textbf{85.6} \textcolor{OliveGreen}{(+3.9)}                 & \textbf{96.4} \textcolor{OliveGreen}{(+1.2)}                 \\
\bottomrule
\end{tabular}
}
\end{table*}

To verify that our method behaves as intended and indeed yields better binding at the visual token level, we visualize \textit{the change in the attention map} when training with our approach compared to SigLIP. \cref{fig:attention_change} shows results for the caption ``black sweater'' and ``green tablecloth'' next to the corresponding images.
For instance, it can be seen in \cref{subfig:black_sweater} that the sign in the background receives considerably less attention from our C\textsuperscript{2}LIP than from SigLIP. 
At the same time, attention increases or remains high for most patches that cover the sweater. 
This clearly illustrates that training with our model improves binding at the patch level (i.e., before global pooling).
Moreover, the quantitative results presented in Sec.~\ref{ssec:quantitative_evaluation} further demonstrate that binding improves after global pooling.

\subsection{Ablation studies}\label{ssec:ablations}

We perform an ablation study of the components of our pipeline and report the results in \cref{tab:ablation}.
As can be seen, we observe large improvements across the benchmark when using the noun-phrase loss ($L_{npc}$).
Adding the cross-modal attention pooling loss ($L_{xac}$) yields additional gains, particularly for the ``Swap'' and ``Add'' tasks.
These results demonstrate that focusing on positive examples, i.e., using concept-centric text tokens as extra positives, provides a substantial boost for compositionality.
When a cross-modal attention pooling mechanism is also employed, the network can learn more easily, encouraging stronger binding before the global pooling stage.
Consequently, performance improves consistently on all benchmarks.

\subsection{C\textsuperscript{2}LIP improves visual instruction tuning}

In our compositionality experiments, C\textsuperscript{2}LIP demonstrates consistent superior accuracies in comparison to similar-sized baselines. However, these experiments are limited within the cross-modal retrieval task, and do not provide insights on whether these enhanced capabilities could influence other downstream applications, such as using C\textsuperscript{2}LIP as vision encoder for an LLM to train a vision-language model (VLM). Thus, we conduct experiments based on the LLaVa~\cite{llava_15} instruction tuning framework, where the frozen image encoder is combined with an LLM for image-to-text generation. In particular, we replace the CLIP ViT image encoder of LLaVa with SigLIP and C\textsuperscript{2}LIP vision encoders, and train the adapter MLP and LLM following the 2-stage recipe of LLaVa on the same data, resulting in two VLMs, LLaVA-SigLIP and LLaVA-C\textsuperscript{2}LIP, respectively. The resulting models are evaluated on SugarCrepe and SugarCrepe++, where the cosine similarity score is substituted with the VQA image-text matching score (VQAScore)~\cite{vqascore}. The performance metrics are summarized in \cref{tab:lvlm}. LLaVA-C\textsuperscript{2}LIP outperforms LLaVA-SigLIP on both SugarCrepe and SugarCrepe++ by 0.4\% and 1.6\%, respectively, showing that the enhanced compositionality comprehension capabilities of C\textsuperscript{2}LIP also transfers to the VLM that utilizes its vision encoder.

\begin{table*}[!ht]
\centering
\caption{\textbf{Performance of VLM using C\textsuperscript{2}LIP vision encoder.} We follow the LLaVa~\cite{llava_15} recipe to train VLMs with frozen SigLIP and C\textsuperscript{2}LIP image encoders, resulting in LLaVA-SigLIP and LLaVA-C\textsuperscript{2}LIP, respectively. LLaVA-C\textsuperscript{2}LIP shows improvements on compositionality, demonstrating that the vision encoder trained with our proposed concept-centric approach also benefits the VLM that utilizes it.}
\label{tab:lvlm}
\small
\begin{tabular}{l|ccc|c|cc|c}
\toprule
\multicolumn{1}{c|}{\multirow{2}{*}{Model}} & \multicolumn{4}{c|}{SugarCrepe}                                                                              & \multicolumn{3}{c}{SugarCrepe++}                                                  \\
\multicolumn{1}{c|}{}                       & \multicolumn{1}{c}{Replace} & \multicolumn{1}{c}{Swap} & \multicolumn{1}{c}{Add} & \multicolumn{1}{c|}{\textbf{Average}} & \multicolumn{1}{c}{Replace} & \multicolumn{1}{c}{Swap} & \multicolumn{1}{c}{\textbf{Average}} \\
\midrule
SigLIP-ViT-B-16                            & 84.1                        & 65.8                     & 86.5                    & 79.5                     & 73.8                        & 48.0                     & 63.5                     \\
C\textsuperscript{2}LIP                                      & \textbf{89.1}               & 75.2                     & \textbf{93.3}           & \textbf{86.3}            & \textbf{79.7}               & 55.2                     & \textbf{69.9}            \\
\midrule
LLaVA-SigLIP                               & 86.9                        & 77.0                     & 85.0                    & 83.5                     & 76.5                        & 55.4                     & 68.1                     \\
LLaVA-C\textsuperscript{2}LIP                                & 86.8                        & \textbf{78.5}            & 85.0                    & 83.9                     & 77.6                        & \textbf{57.8}            & 69.7                     \\
\bottomrule
\end{tabular}
\end{table*}

\section{Conclusion}

We propose a new method to improve the compositionality of CLIP-like models.  
We process the training captions to 
extract noun-phrases, i.e., short segments that correspond to a single object and its attributes. 
Using the caption decomposition we obtain a context embedding for each concept in the caption. 
By aligning each concept embedding with the global image embedding and subsequently using a parameter-free attention-pooling function to create visual-domain concept embeddings, we enforce alignment between visual and textual concept embeddings via a contrastive loss.
Notably, our method \textbf{C\textsuperscript{2}LIP} does not rely on any hard negatives, which often yield improvements that fail to generalize to slightly different compositionality tasks. 
Our approach is simple and incurs no additional computational overhead at inference time; in fact, inference remains identical to the original
model.
We evaluate \textbf{C\textsuperscript{2}LIP} on standard compositionality benchmarks as well as a variety of zero-shot and retrieval tasks. \textbf{C\textsuperscript{2}LIP} achieves consistent improvements on all tested compositionality benchmarks and on most standard retrieval tasks. Owing to its consistently strong performance, \textbf{C\textsuperscript{2}LIP} attains overall SOTA performance.

\textbf{Limitations and future work.}
While our approach leads to consistent improvements without sacrificing general capabilities, we do observe a small drop in performance on ImageNet. Future work should explore whether it is possible to reduce this loss even further.
Similarly, our approach clearly improves the compositional representation, but the final pooling layer remains suboptimal. Future work should explore pooling operations that naturally preserve compositionality,  and the extensibility of our proposed method to other types of compositional reasoning beyond object-attribute binding.

{
    \small
    \bibliographystyle{ieeenat_fullname}
    \bibliography{main}
}

\appendix

\setcounter{table}{0}
\renewcommand{\thetable}{A\arabic{table}}

\setcounter{section}{0}
\renewcommand{\thesection}{\Alph{section}}

\section{Additional experimental results}

We provide additional results of our method, C\textsuperscript{2}LIP, and the baseline contrastive models employing the same ViT-B backbone listed in \cref{a_tab:baselines}. Please refer to the main paper for details of the benchmarks and evaluation protocol.

This section is organized as follows. First, we show that our proposed loss function is not sensitive to scaling hyperparameters in \cref{a_ssec:loss_hyperparam}. Next, the evaluation results on compositionality benchmarks including SugarCrepe and SugarCrepe++ are described in \cref{a_ssec:composition}. Furthermore, we discuss zero-shot retrieval performance in \cref{a_ssec:retrieval}. Finally, the zero-shot classification results are given in \cref{a_ssec:classification}.

\begin{table}[ht]
\centering
\caption{\textbf{Baseline methods.} Summary of baseline methods and their corresponding training datasets. The last column indicates whether the provided checkpoints were trained from scratch (\checkmark).}
\label{a_tab:baselines}
\resizebox{\columnwidth}{!}{
\begin{tabular}{l|c|c}
\toprule
\multicolumn{1}{c|}{Model}               & Training data          & Train from scratch \\\midrule
\textit{\textbf{Composition-aware}}     & \multicolumn{1}{l|}{}   &                    \\
\hspace{1em}CE-CLIP~\cite{ce-clip-Zhang2023-vj}                                 & MSCOCO                   &                    \\
\hspace{1em}NegCLIP~\cite{negclip_iclr23}                                 & MSCOCO                   &                    \\
\hspace{1em}CLIC~\cite{clic_Peleg2025}                                    & LAION-1.5B             &                    \\\cmidrule{2-2}
\hspace{1em}DAC~\cite{Doveh2023-qp}                                     & \multirow{10}{*}{CC3M} &                    \\
\hspace{1em}SLVC~\cite{Doveh2022-yg}                                    &                        &                    \\
\hspace{1em}CoN-CLIP~\cite{Singh2024-cn}                                &                        &                    \\
\hspace{1em}TripletCLIP~\cite{Patel2024-qj}                             &                        & \checkmark                  \\
\textit{\textbf{Codebook-based}}        &                        &                    \\
\hspace{1em}Codebook-CLIP~\cite{codebook-clip-chen2023}                           &                        & \checkmark                  \\
\hspace{1em}IL-CLIP~\cite{il-clip-zheng2024iterated}                                 &                        & \checkmark                  \\
\textit{\textbf{Fine-grained training}} &                        &                    \\
\hspace{1em}DreamLIP-3m~\cite{dreamlip-zheng}                             &                        & \checkmark                  \\
\hspace{1em}FLAIR-3m~\cite{xiao2025flair}                                &                        & \checkmark                  \\\cmidrule{2-2}
\hspace{1em}FG-CLIP~\cite{fg-clip-Xie2025}                                 & LAION-2B + FineHARD    & \checkmark                  \\
\hspace{1em}FineCLIP~\cite{fineclip-dong}                                & MSCOCO                   &                    \\
\hspace{1em}LLIP~\cite{llip}                                            & Common Crawl 12.8B        & \checkmark        \\
\bottomrule
\end{tabular}
}
\end{table}

\subsection{Sensitivity to hyperparameters in objective function}
\label{a_ssec:loss_hyperparam}

Tab~\ref{a_tab:sensitivity} shows the average accuracies on SugarCrepe of models trained with different values of $\lambda_{hnc}$ \& $\lambda_{xac}$. The variation in performance scores is marginal, showing that our proposed loss function function is robust to non-optimal values of of these hyperparameters. We selected the combination of $(1, 0.01)$ in all our experiments.

\begin{table}[!ht]
\centering
\caption{\textbf{Ablation of trade-off hyperparameters in the objective function.} We experimented with different values of $\lambda_{hnc}$ \& $\lambda_{xac}$ showing our proposed method incurs little sensitivity to these hyperparameters.}
\label{a_tab:sensitivity}
\begin{tabular}{c|c|c}
\hline
$\lambda_{hnc}$ & $\lambda_{xac}$ & SugarCrepe Average Accuracy \\
\hline
0.5                                                 & 0.5                                                 & 84.6                        \\
0.5                                                 & 0.1                                                 & 85.2                        \\
0.5                                                 & 0.01                                                & 85.2                        \\
1                                                   & 0.5                                                 & 85.1                        \\
1                                                   & 0.01                                                & \textbf{85.6}               \\
\hline
\end{tabular}
\end{table}

\subsection{Compositionality evaluation}
\label{a_ssec:composition}

In addition to the task-specific average scores shown in Tab. 2 of the main paper, we include all accuracy scores of all competing methods on all sub-tasks of SugarCrepe and SugarCrepe++ benchmarks in \cref{a_tab:sugarcrepe} and \cref{a_tab:sugarcrepe_pp}, respectively.

\vspace{\baselineskip}
\noindent\textbf{SugarCrepe benchmark.} As can be seen in \cref{a_tab:sugarcrepe}, our concept centric contrastive learning method improves the performance of the original base model, SigLIP, by $7.67\%$ on average, surpassing most other composition-aware methods as the second best performing model, only behind DAC-LLM~\citep{Doveh2023-qp} by 0.8 percentage points. Notably, these methods rely on training with hard-negatives to induce the compositionality representations. Instead our method emphasizes better exploiting regular data, with auxiliary concept centric objectives, to improve compositional representations. In particular, C\textsuperscript{2}LIP excels at recognizing incorrect objects in the image, evidenced by the highest scores on ``Replace Object'' and ``Add Object'' sub-tasks. Moreover, C\textsuperscript{2}LIP exhibits strong attribute-binding capabilities, with the highest score on ``Swap Attribute'' and second best on ``Replace Attribute''. Our method, however, lags behind in ``Replace Relation'' and ``Swap Object'' sub-tasks. On the other hand, the substantial improvements on these two sub-tasks compared to the original SigLIP model demonstrate the effectiveness of our method.
The results on SugarCrepe should be interpreted carefully, as the benchmark is insufficient to evaluate lexical sensitivity and semantic understanding \citep{sugarcrepe++}.

\vspace{\baselineskip}
\noindent\textbf{SugarCrepe++ benchmark.} SugarCrepe++ \citep{sugarcrepe++} resolves this problem by extending the protocol of SugarCrepe by using two positive captions and requiring both of them to be higher ranked than the negative, to be considered correct. By that, it aims to address the limitation of SugarCrepe, where the caption patterns can be, to some extent, imitated to create custom training data. However, the second positive captions in SugarCrepe++ aim to evaluate the generalization capabilities of contrastive models. Intuitively, the concepts in the second caption remain the same as in the first, described differently, thus a model trained to fit the pattern in the first caption may no longer align to the second. As shown in \cref{a_tab:sugarcrepe_pp}, the methods relying on custom hard-negative training data incur a substantial performance drop across SugarCrepe++ tasks. In contrast, our method performs consistently well across all tasks, showcasing both effective compositional representation as well as strong generalization capabilities. On average C\textsuperscript{2}LIP outperforms the baselines by a large margin, made possible by our proposed concept centric learning framework.

\begin{table*}[ht]
\centering
\caption{\textbf{SugarCrepe compositionality benchmark -- all subtasks}. The results in this table are summarized in Tab.~2 of the main paper. We compare the accuracy of C\textsuperscript{2}LIP with baseline methods on different tasks. On average, C\textsuperscript{2}LIP is the second best method, only 0.8 percentage points below DAC-LLM, while being trained on much less data, without custom compositional captions.}
\label{a_tab:sugarcrepe}
\begin{tabular}{l|cccc|ccc|ccc|c}
\toprule
\multicolumn{1}{c|}{\multirow{2}{*}{Models}} & \multicolumn{4}{c|}{Replace}                                                                                         & \multicolumn{3}{c|}{Swap}                                                             & \multicolumn{3}{c|}{Add}                                                              & \multicolumn{1}{c}{\multirow{2}{*}{\textbf{Average}}} \\
\cmidrule(l{1pt}r{1pt}){2-5}\cmidrule(l{1pt}r{1pt}){6-8}\cmidrule(l{1pt}r{1pt}){9-11}
\multicolumn{1}{c|}{}                        & \multicolumn{1}{c}{Obj} & \multicolumn{1}{c}{Attr} & \multicolumn{1}{c}{Rel} & \multicolumn{1}{c|}{Avg} & \multicolumn{1}{c}{Obj} & \multicolumn{1}{c}{Attr} & \multicolumn{1}{c|}{Avg} & \multicolumn{1}{c}{Obj} & \multicolumn{1}{c}{Attr} & \multicolumn{1}{c|}{Avg} & \multicolumn{1}{c}{}                             \\
\midrule
SigLIP ViT-B/16                             & 95.3                       & 86.7                          & 70.3                         & 84.1                    & 60.0                       & 71.5                          & 65.8                    & 89.1                       & 83.8                          & 86.5                    & 79.5                                             \\
CLIP (OpenAI) ViT-B/32                      & 90.9                       & 80.0                          & 69.2                         & 80.0                    & 61.2                       & 64.1                          & 62.7                    & 77.2                       & 68.8                          & 73.0                    & 73.1                                             \\
CLIP (OpenAI) ViT-B/16                      & 93.5                       & 81.1                          & 66.7                         & 80.4                    & 60.0                       & 65.0                          & 62.5                    & 78.5                       & 66.9                          & 72.7                    & 73.1                                             \\
\midrule
FG-CLIP                                     & {\ul 95.9}                 & 87.1                          & 72.2                         & 85.1                    & 66.5                       & 73.3                          & 69.9                    & 87.6                       & 81.8                          & 84.7                    & 80.6                                             \\
FineCLIP                                    & 95.6                       & 85.2                          & 75.0                         & 85.3                    & 63.3                       & 70.3                          & 66.8                    & 90.0                       & 80.8                          & 85.4                    & 80.0                                             \\
DreamLIP-3m                                 & 87.2                       & 77.3                          & 68.1                         & 77.5                    & 56.3                       & 72.1                          & 64.2                    & 74.3                       & 71.5                          & 72.9                    & 72.4                                             \\
FLAIR-3m                                    & 91.4                       & 82.3                          & 70.5                         & 81.4                    & 63.2                       & 78.5                          & 70.9                    & 84.5                       & 76.6                          & 80.6                    & 78.1                                             \\
LLIP                    & 89.6      & 79.4      & 67.6      & 78.9      & 55.9      & 59.6      & 57.8      & 79.1      & 63.7      & 71.4      & 70.7 \\
\midrule
Codebook-CLIP                               & 53.5                       & 51.4                          & 58.7                         & 54.5                    & 45.7                       & 49.2                          & 47.5                    & 57.3                       & 43.8                          & 50.6                    & 51.4                                             \\
IL-CLIP                                     & 53.1                       & 54.1                          & 51.1                         & 52.8                    & 57.6                       & 52.7                          & 55.2                    & 54.9                       & 48.6                          & 51.8                    & 53.2                                             \\
\midrule
CE-CLIP                                     & 93.1                       & 88.8                          & 79.0                         & 87.0                    & 72.8                       & {\ul 77.0}                    & {\ul 74.9}              & {\ul 92.4}                 & 93.4                          & 92.9                    & 85.2                                             \\
NegCLIP                                     & 92.7                       & 85.9                          & 76.5                         & 85.0                    & \textbf{75.2}              & 75.4                          & \textbf{75.3}           & 88.8                       & 82.8                          & 85.8                    & 82.5                                             \\
CLIC                                        & 95.6                       & 86.6                          & 75.3                         & 85.8                    & 71.0                       & 74.0                          & 72.5                    & 88.4                       & 91.2                          & 89.8                    & 83.1                                             \\
DAC-SAM                                     & 91.2                       & 85.9                          & {\ul 83.9}                   & 87.0                    & 71.8                       & 75.1                          & 73.5                    & 87.5                       & {\ul 95.7}                    & 91.6                    & 84.4                                             \\
DAC-LLM                                     & 94.5                       & \textbf{89.5}                 & \textbf{84.4}                & \textbf{89.5}           & {\ul 75.1}                 & 74.2                          & 74.6                    & 89.7                       & \textbf{97.7}                 & {\ul 93.7}              & \textbf{86.4}                                    \\
SLVC-R                                      & 91.3                       & 81.4                          & 64.1                         & 78.9                    & 68.6                       & 69.1                          & 68.8                    & 79.5                       & 91.3                          & 85.4                    & 77.9                                             \\
SLVC-RL                                     & 88.1                       & 76.8                          & 62.7                         & 75.9                    & 64.5                       & 66.5                          & 65.5                    & 75.8                       & 81.2                          & 78.5                    & 73.7                                             \\
CoN-CLIP                                    & 92.5                       & 79.7                          & 60.1                         & 77.4                    & 58.8                       & 66.1                          & 62.4                    & 86.7                       & 78.2                          & 82.4                    & 74.6                                             \\
TripletCLIP                                 & 94.4                       & 85.7                          & 80.9                         & 87.0                    & 70.2                       & 69.7                          & 69.9                    & 90.4                       & 86.1                          & 88.3                    & 82.5                                             \\
\midrule
SigLIP ViT-B/16 (ft. CC3m)                  & 95.8                       & 87.4                          & 73.5                         & 85.6                    & 65.3                       & 74.2                          & 69.7                    & 88.9                       & 87.0                          & 87.9                    & 81.7                                             \\
C\textsuperscript{2}LIP                                        & \textbf{96.7}              & {\ul 89.2}                    & 78.9                         & {\ul 88.3}              & 67.3                       & \textbf{78.8}                 & 73.1                    & \textbf{93.5}              & 94.9                          & \textbf{94.2}           & {\ul 85.6}                            \\
\bottomrule
\end{tabular}
\end{table*}

\begin{table*}[ht]
\centering
\caption{\textbf{SugarCrepe++ compositionality benchmark -- all subtasks}. The results in this table are summarized in Tab.~2 of the main paper. We compare the accuracy of C\textsuperscript{2}LIP with baseline methods on different compositionality tasks. SugarCrepe++ addresses the limitation of the SugarCrepe benchmark, where the positive and negative captions are ``hackable'' by using custom training data created using similar rules. Here we observe significant performance drops from the baseline methods, while our model, which maintains generalization capabilities, achieves the best result overall.}
\label{a_tab:sugarcrepe_pp}
\resizebox{\textwidth}{!}{
\begin{tabular}{l|cccc|cccc|ccc|ccc|c}
\toprule
\multicolumn{1}{c|}{\multirow{2}{*}{Models}} & \multicolumn{4}{c|}{Replace - I2T}                                                                                   & \multicolumn{4}{c|}{Replace - TOT}                                                                                   & \multicolumn{3}{c|}{Swap - I2T}                                                       & \multicolumn{3}{c|}{Swap - TOT}                                                       & \multicolumn{1}{c}{\multirow{2}{*}{\textbf{Average}}} \\
\cmidrule(l{1pt}r{1pt}){2-5}\cmidrule(l{1pt}r{1pt}){6-9}\cmidrule(l{1pt}r{1pt}){10-12}\cmidrule(l{1pt}r{1pt}){13-15}
\multicolumn{1}{c|}{}                        & \multicolumn{1}{c}{Obj} & \multicolumn{1}{c}{Attr} & \multicolumn{1}{c}{Rel} & \multicolumn{1}{c|}{Avg} & \multicolumn{1}{c}{Obj} & \multicolumn{1}{c}{Attr} & \multicolumn{1}{c}{Rel} & \multicolumn{1}{c|}{Avg} & \multicolumn{1}{c}{Obj} & \multicolumn{1}{c}{Attr} & \multicolumn{1}{c|}{Avg} & \multicolumn{1}{c}{Obj} & \multicolumn{1}{c}{Attr} & \multicolumn{1}{c|}{Avg} & \multicolumn{1}{c}{}                             \\
\midrule
SigLIP ViT-B/16                             & 91.2                       & 75.5                          & 54.8                         & 73.8                    & 79.2                       & 64.2                          & 45.0                         & 62.8                    & 39.6                       & 56.3                          & 48.0                    & 22.9                       & 46.4                          & 34.7                    & 57.5                                             \\
CLIP (OpenAI) ViT-B/32                      & 86.7                       & 65.6                          & 56.3                         & 69.5                    & 83.7                       & 59.3                          & 38.6                         & 60.5                    & 46.1                       & 45.2                          & 45.7                    & 19.1                       & 35.6                          & 27.4                    & 53.6                                             \\
CLIP (OpenAI) ViT-B/16                      & 89.6                       & 67.6                          & 53.2                         & 70.1                    & 84.4                       & 57.2                          & 39.0                         & 60.2                    & 39.2                       & 48.4                          & 43.8                    & 16.3                       & 31.4                          & 23.9                    & 52.6                                             \\
\midrule
FG-CLIP                                     & {\ul 92.6}                 & {\ul 76.8}                    & 58.0                         & 75.8                    & 90.6                       & 67.8                          & 44.2                         & 67.5                    & 47.8                       & 55.3                          & 51.5                    & \textbf{29.4}              & 47.0                          & 38.2                    & {\ul 60.9}                                       \\
FineCLIP                                    & 90.8                       & 70.7                          & 57.0                         & 72.8                    & 91.3                       & 67.9                          & 47.0                         & 68.7                    & 39.6                       & 48.2                          & 43.9                    & 20.8                       & 32.7                          & 26.8                    & 56.6                                             \\
DreamLIP-3m                                 & 75.8                       & 60.8                          & 46.9                         & 61.2                    & 71.4                       & 50.8                          & 32.4                         & 51.5                    & 34.3                       & 54.4                          & 44.4                    & 20.0                       & 40.1                          & 30.1                    & 48.7                                             \\
FLAIR-3m                                    & 84.3                       & 64.2                          & 50.3                         & 66.3                    & 77.3                       & 58.4                          & 36.8                         & 57.5                    & 40.8                       & 58.3                          & 49.5                    & 24.9                       & 45.5                          & 35.2                    & 54.1                                             \\
LLIP        & 84.1      & 66.0      & 51.9      & 67.3      & 71.2      & 54.8      & 44.5      & 56.8      & 36.7      & 44.9      & 40.8      & 24.5      & 30.2      & 27.3      & 50.9 \\
\midrule
Codebook-CLIP                               & 32.3                       & 32.6                          & 37.8                         & 34.3                    & 18.2                       & 9.8                           & 13.5                         & 13.8                    & 28.2                       & 28.5                          & 28.3                    & 8.6                        & 11.6                          & 10.1                    & 22.1                                             \\
IL-CLIP                                     & 38.0                       & 32.9                          & 32.4                         & 34.4                    & 55.8                       & 18.5                          & 21.3                         & 31.8                    & 39.2                       & 34.7                          & 36.9                    & 9.4                        & 20.0                          & 14.7                    & 30.2                                             \\
\midrule
CE-CLIP                           & 71.9                       & 52.0                          & 45.5                         & 56.5                    & 86.3                       & 64.2                          & 50.5                         & 67.0                    & 36.3                       & 32.3                          & 34.3                    & {\ul 28.6}                 & 36.3                          & 32.5                    & 50.4                                             \\
NegCLIP                           & 87.0                       & 67.1                          & 53.1                         & 69.1                    & \textbf{93.3}              & 70.7                          & 48.6                         & {\ul 70.9}              & {\ul 51.8}                 & 55.0                          & 53.4                    & 27.8                       & {\ul 50.3}                    & {\ul 39.1}              & 60.5                                             \\
CLIC                                        & 91.6                       & 75.9                          & {\ul 62.3}                   & {\ul 76.6}              & 84.7                       & 52.4                          & 37.3                         & 58.1                    & \textbf{55.9}              & {\ul 62.2}                    & \textbf{59.0}           & 22.9                       & 31.2                          & 27.0                    & 57.6                                             \\
DAC-SAM                                     & 64.3                       & 44.3                          & 48.7                         & 52.4                    & 75.9                       & 56.0                          & 48.7                         & 60.2                    & 27.8                       & 33.5                          & 30.6                    & 11.4                       & 25.4                          & 18.4                    & 43.6                                             \\
DAC-LLM                                     & 65.7                       & 47.7                          & 47.6                         & 53.7                    & 76.8                       & 59.5                          & 42.3                         & 59.6                    & 31.4                       & 32.9                          & 32.2                    & 11.4                       & 24.8                          & 18.1                    & 44.0                                             \\
SLVC-R                                      & 82.9                       & 61.6                          & 47.7                         & 64.0                    & 89.5                       & 67.4                          & 47.7                         & 68.2                    & 49.4                       & 53.2                          & 51.3                    & 20.4                       & 36.3                          & 28.4                    & 55.6                                             \\
SLVC-RL                                     & 81.0                       & 57.1                          & 47.5                         & 61.9                    & {\ul 91.6}                 & 67.0                          & 51.3                         & 70.0                    & 43.3                       & 48.8                          & 46.0                    & 18.4                       & 34.3                          & 26.3                    & 54.0                                             \\
CoN\_CLIP                                   & 87.9                       & 68.1                          & 48.2                         & 68.1                    & 91.5                       & 64.7                          & {\ul 53.9}                   & 70.1                    & 40.0                       & 50.3                          & 45.2                    & 18.8                       & 37.2                          & 28.0                    & 56.1                                             \\
TripletCLIP                                 & 84.9                       & 66.0                          & 58.7                         & 69.9                    & 89.0                       & {\ul 71.1}                    & 48.1                         & 69.4                    & 38.4                       & 4.4                           & 21.4                    & 18.8                       & 38.1                          & 28.5                    & 51.7                                             \\
\midrule
SigLIP ViT-B/16 (ft. CC3m)                  & 91.7                       & 74.2                          & 54.6                         & 73.5                    & 85.4                       & 69.3                          & 48.9                         & 67.9                    & 42.5                       & 57.2                          & 49.8                    & 23.7                       & 49.6                          & 36.6                    & 59.7                                             \\
C\textsuperscript{2}LIP                                        & \textbf{93.9}              & \textbf{79.8}                 & \textbf{65.4}                & \textbf{79.7}           & 91.1                       & \textbf{80.2}                 & \textbf{54.7}                & \textbf{75.3}           & 44.1                       & \textbf{66.2}                 & {\ul 55.2}              & {\ul 28.6}                 & \textbf{59.8}                 & \textbf{44.2}           & \textbf{66.4}                \\
\bottomrule
\end{tabular}}
\end{table*}

\subsection{Zero-shot retrieval evaluation}
\label{a_ssec:retrieval}

We evaluate our proposed method and the baselines on two regular retrieval benchmarks: MSCOCO and Flickr30k, and two fine-grained retrieval benchmarks: DOCCI and Image-in-words (IIW). We report Recall@5 scores of image-to-text and text-to-image retrieval tasks, as summarized in \cref{a_tab:retrieval}. As shown in this table, the composition-aware methods incur degraded retrieval performance compared to the base CLIP model. In contrast, C\textsuperscript{2}LIP performs consistently well. It is the best performing model on most tasks, and on par with the top models for the remaining tasks. Our model enjoys 1.9 percentage point improvement over the original SigLIP on average. These results prove that our training method not only effectively maintains, but can also improve the generalization capability of the original model.

\begin{table*}[ht]
\centering
\caption{\textbf{Zero-shot retrieval benchmarks -- all subtasks}. The results in this table are summarized in Tab.~2 of the main paper. We record Recall@5 metrics of all models on text-to=image (t2i) and image-to-text (i2t) tasks, across two standard retrieval benchmarks: MSCOCO and Flickr30K, and two fine-grained retrieval benchmarks: DOCCI and Image-in-words (IIW). It can be observed that C\textsuperscript{2}LIP can maintain or improve the performance of the original base model across all tasks, with the best average score.}
\label{a_tab:retrieval}
\begin{tabular}{l|cc|cc|cc|cc|c}
\toprule
\multicolumn{1}{c|}{\multirow{2}{*}{Models}} & \multicolumn{2}{c|}{MSCOCO}                          & \multicolumn{2}{c|}{Flickr30k}                     & \multicolumn{2}{c|}{DOCCI}                         & \multicolumn{2}{c|}{IIW}                & \multicolumn{1}{c}{\multirow{2}{*}{\textbf{Average}}} \\
\cmidrule(l{1pt}r{1pt}){2-3}\cmidrule(l{1pt}r{1pt}){4-5}\cmidrule(l{1pt}r{1pt}){6-7}\cmidrule(l{1pt}r{1pt}){8-9}
\multicolumn{1}{c|}{}                        & \multicolumn{1}{c}{t2i} & \multicolumn{1}{c|}{i2t} & \multicolumn{1}{c}{t2i} & \multicolumn{1}{c|}{i2t} & \multicolumn{1}{c}{t2i} & \multicolumn{1}{c|}{i2t} & \multicolumn{1}{c}{t2i} & \multicolumn{1}{c|}{i2t} & \multicolumn{1}{c}{}                         \\
\midrule
SigLIP ViT-B/16                             & 72.4                    & 85.4                    & 92.3                    & 98.0                    & 35.8                    & {\ul 82.0}              & 53.0                    & 97.5                    & 77.1                                         \\
CLIP (OpenAI) ViT-B/32                      & 56.0                    & 74.9                    & 83.4                    & 94.6                    & 27.1                    & 67.1                    & 44.7                    & 94.3                    & 67.8                                         \\
CLIP (OpenAI) ViT-B/16                      & 58.4                    & 76.7                    & 85.6                    & 96.2                    & 29.3                    & 71.5                    & 46.7                    & 95.9                    & 70.0                                         \\
\midrule
FG-CLIP                                     & 71.4                    & 85.5                    & {\ul 93.0}              & {\ul 98.6}              & 33.9                    & 79.4                    & 52.5                    & \textbf{98.7}           & 76.6                                         \\
FineCLIP                                    & {\ul 73.7}              & 84.2                    & 90.2                    & 96.7                    & 27.9                    & 61.9                    & 44.6                    & 89.5                    & 71.1                                         \\
DreamLIP-3m                                 & 55.2                    & 67.2                    & 76.6                    & 89.6                    & 39.3                    & 78.2                    & \textbf{58.5}           & 97.4                    & 70.2                                         \\
FLAIR-3m                                    & 65.6                    & 77.3                    & 86.5                    & 94.3                    & 30.9                    & 63.4                    & 53.0                    & 92.7                    & 70.5                                         \\
LLIP        & 62.3      & 72.8      & 87.3      & 93.1      & 28.6      & 65.2      & 48.3      & 93.1      & 68.8 \\
\midrule
Codebook-CLIP                               & 0.1                     & 0.1                     & 0.5                     & 0.6                     & 0.1                     & 0.1                     & 0.8                     & 0.5                     & 0.3                                          \\
IL-CLIP                                     & 0.1                     & 0.1                     & 0.3                     & 0.5                     & 0.1                     & 0.1                     & 0.7                     & 0.7                     & 0.3                                          \\
\midrule
CE-CLIP                                     & 69.5                    & 74.3                    & 86.4                    & 88.4                    & 19.1                    & 42.4                    & 31.4                    & 68.8                    & 60.0                                         \\
NegCLIP                                     & 68.4                    & 79.3                    & 89.5                    & 95.2                    & 26.4                    & 64.0                    & 43.5                    & 89.4                    & 69.5                                         \\
CLIC                                        & 62.9                    & 71.9                    & 88.2                    & 94.0                    & 33.1                    & 69.4                    & 52.2                    & 94.6                    & 70.8                                         \\
DAC-SAM                                     & 59.7                    & 57.9                    & 85.5                    & 82.5                    & 26.9                    & 35.5                    & 45.5                    & 55.7                    & 56.2                                         \\
DAC-LLM                                     & 63.5                    & 54.5                    & 87.8                    & 79.6                    & 24.8                    & 29.4                    & 40.9                    & 46.9                    & 53.4                                         \\
SLVC-R                                      & 62.0                    & 71.7                    & 87.2                    & 93.1                    & 29.9                    & 54.3                    & 48.6                    & 83.7                    & 66.3                                         \\
SLVC-RL                                     & 62.3                    & 71.8                    & 87.4                    & 92.5                    & 29.7                    & 53.9                    & 48.0                    & 86.6                    & 66.5                                         \\
CoN-CLIP                                    & 54.6                    & 67.9                    & 84.2                    & 87.9                    & 27.2                    & 64.9                    & 42.6                    & 91.5                    & 65.1                                         \\
TripletCLIP                                 & 53.3                    & 55.6                    & 80.9                    & 82.6                    & 25.3                    & 54.2                    & 43.0                    & 82.0                    & 59.6                                         \\
\midrule
SigLIP ViT-B/16 (ft. CC3m)                  & {\ul 73.7}              & {\ul 87.0}              & 92.8                    & 98.4                    & {\ul 36.2}              & \textbf{83.3}           & 53.0                    & {\ul 97.9}              & {\ul 77.8}                                   \\
C\textsuperscript{2}LIP                                        & \textbf{77.9}           & \textbf{87.5}           & \textbf{95.2}           & \textbf{98.8}           & \textbf{38.0}           & 81.9                    & {\ul 56.1}              & 96.7                    & \textbf{79.0}                               \\
\bottomrule
\end{tabular}
\end{table*}

\subsection{Zero-shot classification evaluation}
\label{a_ssec:classification}

Pretrained contrastive V\&L models are increasingly employed in a variety of downstream tasks in computer vision. One of them is zero-shot classification, which made this class of models popular in the first place. We evaluate our method and all baselines on 11 classification benchmarks, their accuracies are summarized in \cref{a_tab:classification}. Among the composition-aware baselines, all methods significantly drop their performance on zero-shot classification. Only CLIC almost reaches the accuracy of the original CLIP model ($68.1$ and $69$, respectively), thanks to the generated training data that helps improve generalization in addition to the hard-negatives. 
Our model incurs a slight $2.9\%$ performance drop compared to the original SigLIP, which is comparatively small in comparison to the other fine-tuned methods.
The drop in performance is not surprising, as our model was fine-tuned with scene centric objectives and data, whereas the classification tasks are object centric. 
Despite the distributional shift, C\textsuperscript{2}LIP can still retain most zero-shot performance.

\begin{table*}[ht]
\centering
\caption{\textbf{Zero-shot classification benchmarks}. We evaluate classification accuracies of C\textsuperscript{2}LIP and the baselines on 11 standard zero-shot classification benchmarks. We observe significantly better performance of our model compared the composition-aware baselines in general, particularly on the challenging \textit{Cars} and \textit{Aircraft} benchmarks where the classes are actually sub-classes of the same object category.}
\label{a_tab:classification}
\begin{tabular}{l|ccccccccccc|c}
\toprule
\multicolumn{1}{c|}{Model}  & \rotatebox[]{90}{ImageNet1K} & \rotatebox[]{90}{Food-101} & \rotatebox[]{90}{CIFAR-10} & \rotatebox[]{90}{CIFAR-100} & \rotatebox[]{90}{SUN397} & \rotatebox[]{90}{Cars} & \rotatebox[]{90}{Aircraft} & \rotatebox[]{90}{DTD} & \rotatebox[]{90}{Pets} & \rotatebox[]{90}{Caltech-101} & \rotatebox[]{90}{Flowers} & \rotatebox[]{90}{\textbf{Average}} \\
\midrule
SigLIP ViT-B/16            & \textbf{76.1}                  & \textbf{91.6}                & 92.3                         & 72.2                          & 69.9                       & \textbf{90.9}            & {\ul 43.8}                   & {\ul 64.7}              & {\ul 94.1}               & \textbf{88.0}                   & {\ul 86.0}                  & {\ul 79.0}                  \\
CLIP (OpenAI) ViT-B/32     & 63.3                           & 83.9                         & 89.8                         & 64.3                          & 63.2                       & 59.7                     & 19.6                         & 44.0                    & 87.5                     & 83.8                            & 66.5                        & 66.0                        \\
CLIP (OpenAI) ViT-B/16     & 68.4                           & 88.8                         & 90.8                         & 67.0                          & 65.5                       & 64.7                     & 24.5                         & 45.2                    & 89.2                     & 84.0                            & 71.5                        & 69.0                        \\
\midrule
FG-CLIP                    & 69.0                           & 85.2                         & 93.9                   & \textbf{76.4}                 & \textbf{71.2}              & 84.2                     & 24.4                         & 57.2                    & 90.5                     & 86.2                            & 70.1                        & 73.5                        \\
FineCLIP                   & 55.8                           & 60.1                         & {\ul 94.3}                & 69.0                          & 55.9                       & 6.3                      & 10.7                         & 41.6                    & 57.9                     & 85.4                            & 41.4                        & 52.6                        \\
DreamLIP-3m                & 31.6                           & 23.6                         & 75.7                         & 43.7                          & 41.3                       & 3.5                      & 1.6                          & 18.8                    & 28.9                     & 70.3                            & 18.5                        & 32.5                        \\
FLAIR-3m                   & 33.7                           & 24.8                         & 81.9                         & 51.6                          & 47.0                       & 4.0                      & 2.0                          & 24.3                    & 35.8                     & 72.4                            & 20.4                        & 36.2                        \\
LLIP        & 60.8      & 80.5      & \textbf{94.4}      & 69.8      & 57.7      & 77.1      & 22.3      & 53.3      & 78.8      & 85.3      & 42.9      & 65.7 \\
\midrule
Codebook-CLIP              & 0.1                            & 1.0                          & 11.0                         & 1.1                           & 0.4                        & 0.5                      & 1.1                          & 2.0                     & 2.2                      & 0.9                             & 1.2                         & 1.9                         \\
IL-CLIP                    & 0.1                            & 1.0                          & 10.0                         & 1.0                           & 0.1                        & 0.5                      & 1.0                          & 1.6                     & 2.7                      & 0.9                             & 1.1                         & 1.8                         \\
\midrule
CE-CLIP                    & 40.4                           & 59.8                         & 81.2                         & 55.0                          & 44.0                       & 26.1                     & 9.2                          & 28.5                    & 60.9                     & 76.0                            & 37.3                        & 47.1                        \\
NegCLIP                    & 55.7                           & 74.1                         & 85.9                         & 60.9                          & 55.9                       & 46.0                     & 11.8                         & 39.1                    & 82.3                     & 82.5                            & 58.0                        & 59.3                        \\
CLIC                       & 66.6                           & {\ul 88.9}                   & 91.1                         & 68.3                          & 64.0                       & 60.8                     & 23.7                         & 46.7                    & 88.0                     & 84.0                            & 67.5                        & 68.1                        \\
DAC-SAM                    & 52.3                           & 72.3                         & 89.9                         & 63.7                          & 51.4                       & 39.8                     & 9.0                          & 40.2                    & 77.0                     & 78.0                            & 54.2                        & 57.1                        \\
DAC-LLM                    & 51.1                           & 74.5                         & 90.4                         & 63.9                          & 52.1                       & 39.5                     & 11.3                         & 38.5                    & 74.9                     & 79.8                            & 54.9                        & 57.3                        \\
SLVC-R                     & 58.5                           & 81.3                         & 92.3                         & 66.0                          & 62.6                       & 49.6                     & 14.9                         & 39.4                    & 85.8                     & 81.9                            & 59.7                        & 62.9                        \\
SLVC-RL                    & 59.8                           & 81.7                         & 92.0                         & 66.7                          & 63.7                       & 50.6                     & 14.5                         & 39.8                    & 85.0                     & 82.7                            & 61.3                        & 63.4                        \\
CoN-CLIP                   & 63.7                           & 84.5                         & 88.7                         & 63.0                          & 64.0                       & 55.5                     & 19.0                         & 40.4                    & 85.2                     & 83.3                            & 63.3                        & 64.6                        \\
TripletCLIP                & 45.9                           & 58.7                         & 86.9                         & 56.6                          & 53.6                       & 11.7                     & 7.9                          & 33.1                    & 55.2                     & 78.3                            & 45.2                        & 48.5                        \\
\midrule
SigLIP ViT-B/16 (ft. CC3m) & {\ul 75.9}                     & \textbf{91.6}                & 92.4                         & {\ul 72.7}                    & {\ul 70.3}                 & {\ul 90.8}               & \textbf{44.4}                & \textbf{65.1}           & \textbf{94.4}            & \textbf{88.0}                   & \textbf{86.1}               & \textbf{79.2}               \\
C\textsuperscript{2}LIP                       & 73.5                           & 88.7                         & 92.7                         & 72.6                          & 68.1                       & 87.4                     & 32.8                         & \textbf{65.1}           & 92.3                     & {\ul 87.2}                      & 82.9                        & 76.7         \\
\bottomrule
\end{tabular}
\end{table*}

\end{document}